\documentclass{article}

\usepackage{PRIMEarxiv}

\usepackage[utf8]{inputenc} 
\usepackage[T1]{fontenc}    
\usepackage{hyperref}       
\usepackage{url}            
\usepackage{booktabs}       
\usepackage{amsfonts}       
\usepackage{nicefrac}       
\usepackage{microtype}      
\usepackage{lipsum}
\usepackage{fancyhdr}       
\usepackage{graphicx}       
\graphicspath{{media/}}     

\usepackage{multicol}
\usepackage{multirow}
\usepackage{subfig}
\usepackage{times}
\usepackage{soul}
\usepackage{url}
 \usepackage{xcolor,colortbl}

\usepackage{url}
\usepackage{amssymb}
\usepackage{pifont}
\usepackage{rotating}

\newcolumntype{M}[1]{>{\centering\arraybackslash}m{#1}}

\title{Domain-agnostic and Multi-level Evaluation of Generative Models
}

\author{%
 Girmaw Abebe Tadesse \\
 IBM Research - Africa \\
 \texttt{girmaw.abebe.tadesse@ibm.com } \\
 \And
 Jannis Born \\
 IBM Research - Zurich \\
\texttt{jab@zurich.ibm.com}\\
 \And 
 Celia Cintas \\
 IBM Research - Africa \\
 \texttt{celia.cintas@ibm.com} \\
 \And
 William Ogallo\\
 IBM Research - Africa \\
 \texttt{william.ogallo@ibm.com } \\
 \And 
 Dmitry Zubarev \\
 IBM Research - Almaden\\
 \texttt{dmitry.zubarev@ibm.com} \\
 \And
Matteo Manica \\
 IBM Research - Zurich \\
 \texttt{tte@zurich.ibm.com} \\
 \And
Komminist Weldemariam \\
 IBM Research - Yorktown Heights\\
 \texttt{kommy@ibm.com}
}

\begin{document}
\maketitle

\begin{abstract}
While the capabilities of generative models heavily improved in different domains (images, text, graphs, molecules, etc.), their evaluation metrics largely remain based on simplified quantities or manual inspection with limited practicality.
To this end, we propose a framework for Multi-level Performance Evaluation of Generative mOdels (MPEGO), which could be employed across different domains. MPEGO aims to quantify generation performance hierarchically, starting from a sub-feature-based low-level evaluation to a global features-based high-level evaluation. 
MPEGO offers great customizability as the employed features are entirely user-driven and can thus be highly domain/problem-specific while being arbitrarily complex (e.g., outcomes of experimental procedures).
We validate MPEGO using multiple generative models across several datasets from the material discovery domain. An ablation study is conducted to study the plausibility of intermediate steps in MPEGO. Results demonstrate that MPEGO provides a flexible, user-driven, and multi-level evaluation framework, with practical insights on the generation quality. The framework, source code, and experiments will be available  at:~\url{https://github.com/GT4SD/mpego}.
\end{abstract}


\section{Introduction}
Machine Learning (ML) methods, particularly generative models, are effective in addressing critical problems across different domains, which includes material sciences. Examples include the design of novel molecules by combining data-driven techniques and domain knowledge to efficiently search the space of all plausible molecules and generate new and valid ones~\cite{bartok2017machine,von2020retrospective,sousa2021generative,hoffman2022optimizing}. Traditional high-throughput wet-lab experiments, physics-based simulations, and bioinformatics tools for the molecular design process heavily depend on human expertise. These processes require significant resource expenditure to propose, synthesize and test new molecules, thereby limiting the exploration space~\cite{lloyd2020high,polishchuk2013estimation,dimasi2016innovation}.

For example, generative models have been applied to facilitate the material discovery process by employing inverse molecular design problem.
This approach transforms the conventional and slow discovery process by mapping the desired set of properties to a set of structures. The generative process is then optimized to encourage the generation of molecules with those selected properties.
Countless approaches have been suggested for such tasks, most prominently VAEs with different sampling techniques~\cite{chenthamarakshan2020cogmol,born2021datadriven,born2021active}), GANs~\cite{mendez2020novo,maziarka2020mol}, diffusion models~\cite{xu2022geodiff}, flow networks~\cite{jain2022biological} and Transformers~\cite{born2022regression}.

Though the generation capability has been tremendously improved recently,
the quantitative  evaluation of these generative models in different domains remains a grand challenge~\cite{coley2020autonomous}.
Some of the reasons include the multi-objective nature of real discovery problems, the intricacy of evaluating relevant features \textit{in-silico}, and the lack of widely accepted domain- and model-agnostic evaluation frameworks. As a result, existing benchmarks and toolkits in material sciences, such as MOSES~\cite{polykovskiy2020molecular} or GuacaMol~\cite{brown2019guacamol} include limited metrics, such as validity and uniqueness, that lack the capacity to evaluate the complex nature of the generation process (e.g., interactions of multiple properties), thereby less effective to provide meaningful insights to subject matter experts (SMEs) to facilitate practical impact.

In this paper, we introduce a Multi-level Performance Evaluation of Generative mOdels (MPEGO) framework (see Fig.~\ref{fig:overview}), which aims to hierarchically characterize and quantify the capability of generative models, using material discovery domain as a use case.  To that end, MPEGO is a model- and domain-agnostic framework, and its core design is derived from two main requirements: representative \textit{examples} (of  training and generated samples)  and  one or multiple \textit{features} (extracted from these samples). Metrics derived from MPEGO are also interpretable and provide multi-level abstractions of the generation process. Specifically, the contributions of this paper are as follows:

\begin{enumerate}
    \item We provide a multi-level and domain-agnostic evaluation of generative models, starting with sub-feature- or feature-based low-level evaluation to global features based high level-evaluation.
    
      \item We devise a generation frequency analysis that aims to identify and characterize subsets of samples generated with extreme frequencies.
    \item We validate MPEGO framework on multiple generative models (e.g., GCPN~\cite{gcpn2018you},  GraphAF~\cite{graphaf2020shi}, MolGX~\cite{takeda2020molecular}
and Regression Transformer~\cite{born2022regression},  and multiple datasets (e.g., ZINC-250K~\cite{irwin2012zinc} and MOSES~\cite{polykovskiy2020molecular} and CIRCA\footnote{https://circa.res.ibm.com/}.
    \item We conduct ablation studies to analyze MPEGO's sensitivity to design choices.

\end{enumerate}


\section{Related Work}\label{sec:related_work}
The state-of-the-art evaluation approaches for generative  models aim to quantify pre-determined requirements, such as diversity and validity, using a variety of metrics.
Frechet ChemNet Distance (FCD)~\cite{preuer2018frechet} is one of such metrics, and it measures the distance between hidden representations drawn from sets of generated and training samples in the material discovery domain,  which is limited in  providing  sub-feature or feature-level evaluation of models. 
GuacaMol~\cite{brown2019guacamol} is one of the early benchmark platforms for new molecule discovery, which aims to evaluate generative models across different tasks, e.g., fidelity and novelty. 
Molecular Sets (MOSES)~\cite{polykovskiy2020molecular} is another benchmarking framework, which  provides training and testing datasets, and a set of metrics to evaluate the quality and diversity of generated structures to standardize training and model comparisons.

Furthermore, automated characterization of subsets of samples, generated with more or less frequency, i.e., generation frequency analysis, also still remains challenging as the focus is more on latent- or feature-based evaluation. 
Overall, the challenges associated with evaluating generative models could be summarized as follows.  
  First,  multiple evaluation metrics are model-dependent. For example, FCD~\cite{preuer2018frechet} depends on latent representation, and  Maximum-mean discrepancy~\cite{gretton2012kernel} is more specifically used to evaluate graph-based generative models.
State-of-the-art metrics also suffer from limited generalizability (across different levels of feature interactions) and interpretability, e.g., by domain experts, which is critical to achieve trustworthy AI solutions~\cite{liang2022advances}.
In addition, existing evaluation metrics are susceptible to potential flaws in predictive models used in goal-oriented or constrained generation. 
Moreover, existing evaluation strategies lack a generic and standalone evaluation metric that combines both distributional metrics (e.g., uniqueness and diversity) and property-based metrics that score single property.. 
The dependency on a single-constraint objective lacks a principled approach to incorporate multiple  target features.
This becomes a significant challenge when a single and inaccurate  evaluation metric is used, which oversimplifies real discovery problems and hence less practical.

\section{Proposed: MPEGO Framework}\label{sec:proposed}
\begin{figure}[t]
    \centering
    \includegraphics[width=0.7\linewidth]{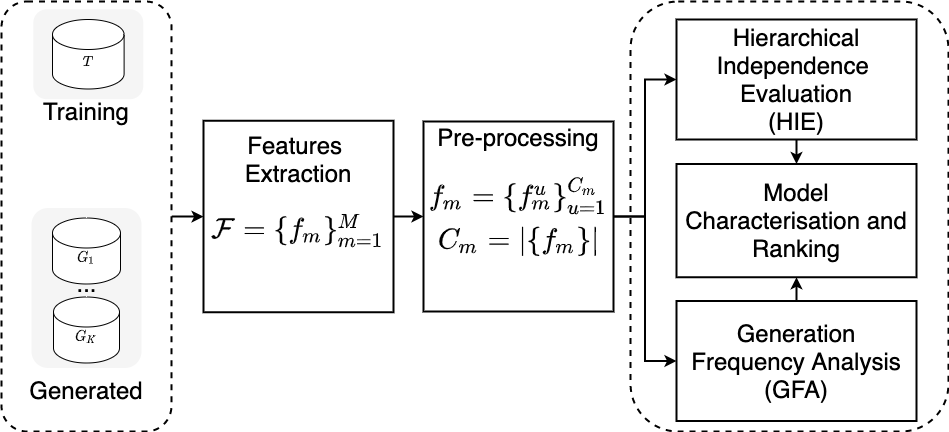}
    \caption{Overview of the MPEGO framework  
} \label{fig:overview}
\end{figure}

The proposed MPEGO framework (see Fig.~\ref{fig:overview}) aims to provide an effective and multi-level characterization of generative models. The multi-level evaluation of MPEGO (see Fig.~\ref{fig:multi-level}) starts from  sub-feature-based low-level evaluation and their step-by-step aggregation to provide high-level evaluations.  
In this section, we first, formulate the critical research questions MPEGO framework is designed to address, followed by the details on its core components.

\subsection{Problem Statement}\label{subsec:problem_formulation}

Let $\mathcal{G}_1, \mathcal{G}_2,\cdots,\mathcal{G}_k,\cdots, \mathcal{G}_K$ be datasets {comprising samples} generated from $K$ black-box generative models ($\Theta_1, \Theta_2, \cdots,\Theta_k, \cdots, \Theta_K$) trained on  a dataset $\mathcal{T}$. Can we evaluate the generation capability of each $\Theta_i$ in a scalable, easily interpretable, and multi-objective manner? 
Specifically, we aim to address two questions. 
\begin{itemize}
    \item[Q1:]Given a set of features characterizing the samples, how do we  quantify the generation capability of each model compared to another model or the training data, based on one or more of these features, i.e., at different levels of abstractions?
    \item[Q2:]What are the characteristics of samples being generated with extreme frequencies (least or most) by each of the generative models, compared with an other model or the training data, i.e., generation frequency analysis? 
\end{itemize}   
To address Q1,  we propose a Hierarchical Independence Evaluation (HIE) that aims to quantify the performance of generative models at different levels of feature interactions hierarchically, starting with a sub-feature level evaluation (e.g., a specific range of a feature) to the global aggregation of multiple features. 
To address Q2, we employ multi-dimensional subset scanning (MDSS)~\cite{neill2013fast} that aims to automatically identify and characterize  over- and under-generated subsets of samples.

\subsection{Feature Extraction and Pre-processing}\label{subsec:feature_extraction_preprocessing}

Given representative examples of generated and training samples, MPEGO starts with the extraction of $M$  features from these samples, $\mathcal{F}=\{f_1,f_2,\cdots,f_m,\cdots,f_M\}$. 
The type of feature values could be binary, {continuous, or categorical, and a further pre-processing could be applied in the follow up steps. For example,  discretization of continuous features is required for sub-feature-level performance evaluation, and  MPEGO provides different  discretization types, e.g., equal width, equal frequency or based on $k$-means.}

\begin{figure*}[t]
    \centering
    \includegraphics[width=1\linewidth]{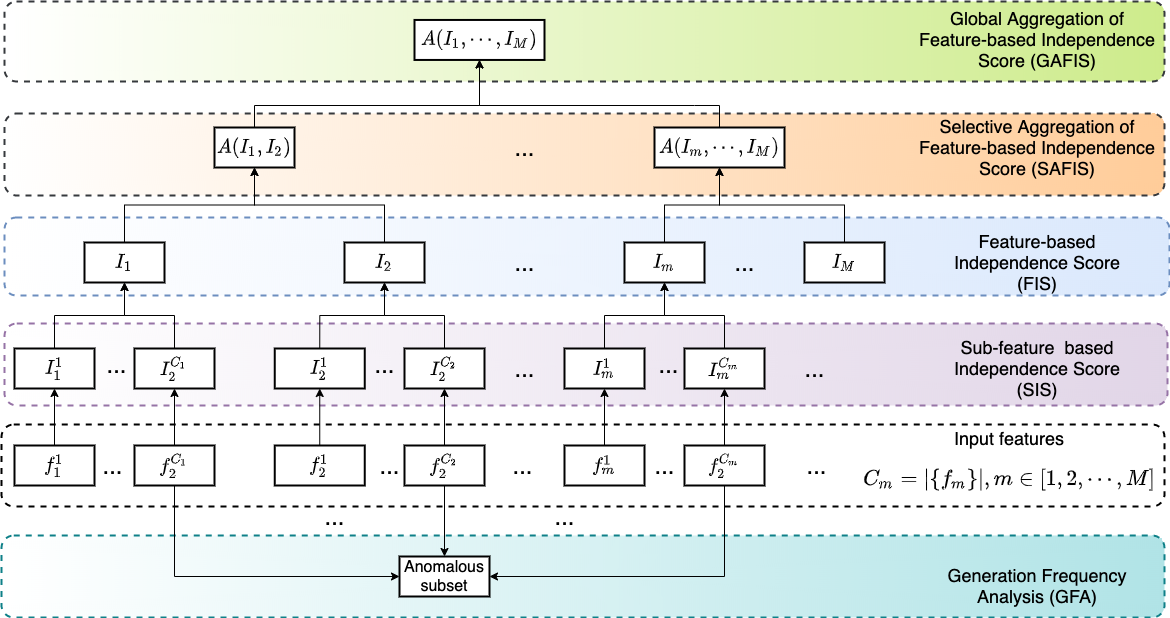}
    \caption{Details of Multi-level evaluation component of the MPEGO framework that comprises Hierarchical Independence Evaluation (i.e., SIS, FIS, SAFIS and GAFIS) and Generation Frequency Analysis (GFA). $A(\cdots)$ represents aggregation operation, and anomalous subset refers to the logical combinations of features that characterize samples generated with extreme frequencies.
    }
    \label{fig:multi-level}
\end{figure*}

\subsection{Hierarchical Independence  Evaluation (HIE)}
HIE follows a bottom-up approach (from a sub-feature to global aggregation levels),   as shown in Fig.~\ref{fig:multi-level}, and it evaluates the performance generative models at  different {levels of feature interactions.}
{The lowest level of evaluation in HIE is the sub-feature level independence score (SIS), which aims to quantify the performance of generative models across different ranges or unique values per feature, e.g., based on a a molecular weight range of $200 - 300$ Daltons. The second layer in HIE represents feature-level independence score (FIS), which quantifies the generation performance for each feature. To this end, FIS could be computed via aggregation of SIS values, thereby providing a weighting strategy for SIS scores per feature. FIS values could also be computed directly, without aggregating SIS values, using a different choice of objective measure is directly applied on the whole feature.The proposed MPEGO framework is flexible to utilize different objective measures (see Appendix D of the Supplementary Material for details). Below we describe the computation of SIS and FIS values, using Yule's Y coefficient~\cite{yule1912methods} as  selected objective measure.} Note that $\mathcal{G}_k$ vs. $\mathcal{T}$ refers to a case where the comparison is between the $kth$ generative model  and the training set $\mathcal{T}$. On the other hand, $\mathcal{G}_k$ vs. $\mathcal{G}_j$ refers to a case when the evaluation between the $jth$ and $kth$ models,  where $j\neq k$. However, we stick with the $\mathcal{G}_k$ vs. $\mathcal{T}$ comparison below in order to ease readability.

 Let   $f_m \in \mathcal{F}$ is  a feature with $C_m$ unique values or ranges, i.e., $f_m =\{f_m^u\}$, $u\in[1,2,\cdots,C_m]$. Note that $C_m$ is the number of unique values for categorical features or the number of bins after discretization of continuous features. 
SIS computation requires the stratification of  both the generated ($\mathcal{G}_k$) and training ($\mathcal{T}$) datasets per each unique value/range $f_m^u$ resulting $\mathcal{G}_{km}^u$  and $\mathcal{T}_{m}^u$, respectively. The complimentary subsets  are then $\widetilde{\mathcal{G}_{km}^u}$  and $\widetilde{\mathcal{T}_{m}^u}$, respectively. Note that  $\widetilde{\mathcal{G}_{km}^u}  =  \mathcal{G}_{km}^u |(f_m \neq f_m^u) =  \mathcal{G}_{k}  - \mathcal{G}_{km}^u$   and $\widetilde{\mathcal{T}_{m}^u}  = \mathcal{T}_{m}^u |(f_m \neq f_m^u) = \mathcal{T}  -\mathcal{T}_{m}^u$.  Accordingly, a $2\times2$ pivot table is generated for each  $f_m^u$ as:
 \begin{center}
 \begin{tabular}{l|c|c}
      & $(f_m=f_m^u)$ &  $(f_m \neq f_m^u)$ \\
\toprule
$\mathcal{G}_{km}$ & $\alpha$ &  $\beta$  \\ \midrule
$\mathcal{T}_m$ & $\delta$  & $\gamma$    \\

\bottomrule
\end{tabular}\label{tab:pivot}
\end{center}
 where $\alpha$ is the number of generated samples in ${\mathcal{G}_{km}^u}$ that are characterized by the  feature value $f_m=f_m^u$,  $\beta$ is the number of generated samples in ${\widetilde{\mathcal{G}_{km}^u}}$ with $f_m \neq f_m^u$. Similarly, $\delta$ and $\gamma$ are the numbers of training samples that satisfy $f_m=f_m^u$ in $\mathcal{T}_m^u$,  respectively. Note that $\alpha + \beta$ are the total numbers of generated samples, i.e., $|{\mathcal{G}_{k}}|$. Similarly,  $\delta + \gamma$ is the number of training samples, i.e., $|\mathcal{T}|$. Then Yule's Y coefficient  is computed from the pivot table as $o_{km}^u \in [-1,1]$:
 \begin{equation}
  o_{km}^u =\frac{\sqrt{P(\mathcal{G}_{km}^u)P(\widetilde{\mathcal{T}_{m}^u})}-\sqrt{P(\widetilde{\mathcal{G}_{km}^u})P(\mathcal{T}_{m}^u)}}{\sqrt{P(\mathcal{G}_{km}^u)P(\widetilde{\mathcal{T}_{m}^u})}+\sqrt{P(\widetilde{\mathcal{G}_{km}^u})P(\mathcal{T}_{m}^u)}} 
  \end{equation}
 \begin{equation}
 o_{km}^u =\frac{\sqrt{\alpha\gamma}-\sqrt{\beta\delta}} {\sqrt{\alpha\gamma}+\sqrt{\beta\delta}}
 \end{equation}
 
 SIS is then computed from $o_{km}^u$ value as $I_{km}^u= 1-|o_{km}^u|$, where $I_{km}^u \in [0,1]$ and higher $I_{km}^u$ reflects higher independence between $\mathcal{G}_{k}$ and $\mathcal{T}$, i.e., SIS $= 1$ represents complete independence.
Feature-level Independence Score (FIS), provides feature-based evaluation, i.e., higher abstraction than SIS. Depending on the objective measure, FIS could be computed as 1) via a weighted aggregation of SIS values, i.e., $I_{km} =\sum_{u=1}^{C_m} \lambda_{km}^u I_{km}^u$  where  $\sum_{u=1}^{C_m} \lambda_{km}^u = 1$ and each $\lambda_{km}^u$ weights  the SIS value of $f_{m}^u$, or 2) via straightforward computation, without using SIS values, when the objective measure is directly applied on each feature without discretization, e.g., using Wasserstein distance. 
{The third layer of HIE is Selective Aggregation of Feature-level Independence Score (SAFIS), which aims to aggregate FIS values from $R<M$ selected features in $\mathcal{F}$. 
For example,  features including scaffolding, fingerprints, aromaticity, and the number of rings could be selected to reflect the structural details of molecules in material discovery domain.  The last layer of HIE is Global Aggregation of Feature-level Independence Score (GAFIS), which is computed via a weighted aggregation of all the FIS values  in $\mathcal{F}$. Note that SAFIS and GAFIS  are computed as 
$\hat{I}=\sum_{r=1}^R \eta_rI_{kr}$, where $\sum_{r=1}^R\eta_r=1$, and $R<M$ for SAFIS and $R=M$ for GAFIS computation}.

\subsection{Generation Frequency Analysis (GFA)}
Generative models trained on the same dataset will hardly generate samples with exact characteristics.
{Thus, there is a potential over- or under-generation of samples with certain characteristics.} To this end, we employ automated stratification of samples using multi-dimensional subset scanning (MDSS)~\cite{neill2013fast,tadesse2022model} to identify subset of samples generated with divergent frequencies. 
Specifically, to identify samples generated with divergent rates by model $\Theta_k$, compared to the training set $\mathcal{T}$, we first merge the corresponding  datasets as $\mathcal{D}=\mathcal{G}_k \cup \mathcal{T}$, and an outcome label ($y$) is generated, such that  $y_i=1$ for a sample in $\mathcal{G}_k$  and $y_i=0$ for a sample in $\mathcal{T}$.  If there are $N_g = |\mathcal{G}_k|$ generated and $N_t = |\mathcal{T}|$ training samples in $\mathcal{D}$, the expectation of generated samples in $\mathcal{D}$ is $e_g= \frac{N_g}{N_g+N_t}$. Thus, GFA aims to identify a group of samples with extreme deviations in their generation rate compared to $e_g$. 
The deviation between the expectation and observation is evaluated by maximizing a Bernoulli likelihood ratio scoring statistic, $\Gamma(\cdot)$. The null hypothesis assumes that the odds of the generated sample in any  subgroup $\mathcal{S}$ is similar to the expected, i.e., $H_0: odds(\mathcal{S})=\frac{e_g}{1-e_g}$; while the alternative hypothesis assumes a constant multiplicative increase in the odds of the generated samples in $\mathcal{S}$, i.e., $H_1: odds(\mathcal{S})=q\frac{e_g}{1-e_g}$ where $q\neq1$. Note that  $q>1$ for over-generated subset, and  $0<q<1$ for under-generated subset. The divergence score for a subgroup ($\mathcal{S}$) with reference $\mathcal{D}$ is formulated as, $\Gamma(\mathcal{S},\mathcal{D})$ and computed as:
 \begin{equation}
\Gamma(\mathcal{S},\mathcal{D}) = \max_q log(q)\sum_{i\in S} y_i - N_s * log(1-e_g + qe_g),
 \end{equation}
 where $N_s$ is the number of samples in $\mathcal{S}$. 
The divergent subset, $\mathcal{S}$, identification is iterated until convergence to a local maximum is found, and the global maximum is subsequently optimized using multiple random restarts.

\section{Experimental Setup}\label{sec:experiment_setup}
\subsection{Training Datasets}
We employed three different datasets in material discovery domain to validate our MPEGO framework.
\paragraph{ZINC-250K}
We utilize the publicly available ZINC-250K\footnote{https://www.kaggle.com/datasets/basu369victor/zinc250k} dataset, which contains $249,455$ small molecules in Simplified Molecular-Input Line-Entry System (SMILES) representation.  Details on ZINC tool are available in ~\cite{irwin2012zinc}.

\paragraph{MOSES}
The benchmark platform MOSES~\cite{polykovskiy2020molecular}, besides implementing popular molecular generation models and metrics, MOSES contains a refined dataset from ZINC\footnote{https://zinc.docking.org/}. The dataset has approximately 2M molecules in total, filtered by certain parameters such as molecular weight ranges, and number of rotatable bonds, among others. 

\paragraph{CIRCA}
We used IBM's Chemical Information Resources for Cognitive Analytics (CIRCA) platform\footnote{https://circa.res.ibm.com/} to construct a dataset of organic salts relevant to production of semiconductors via photolitography with chemical amplification. 
Finally, we were able to evaluate environmental and toxicological properties of 866 anions, that comprise the dataset used in this study. Steps conducted to obtain final version of the dataset could be found in Appendix A in the Supplementary Material. 

     \begin{figure}[t]
    \centering
    {\includegraphics[width=1.0\linewidth]{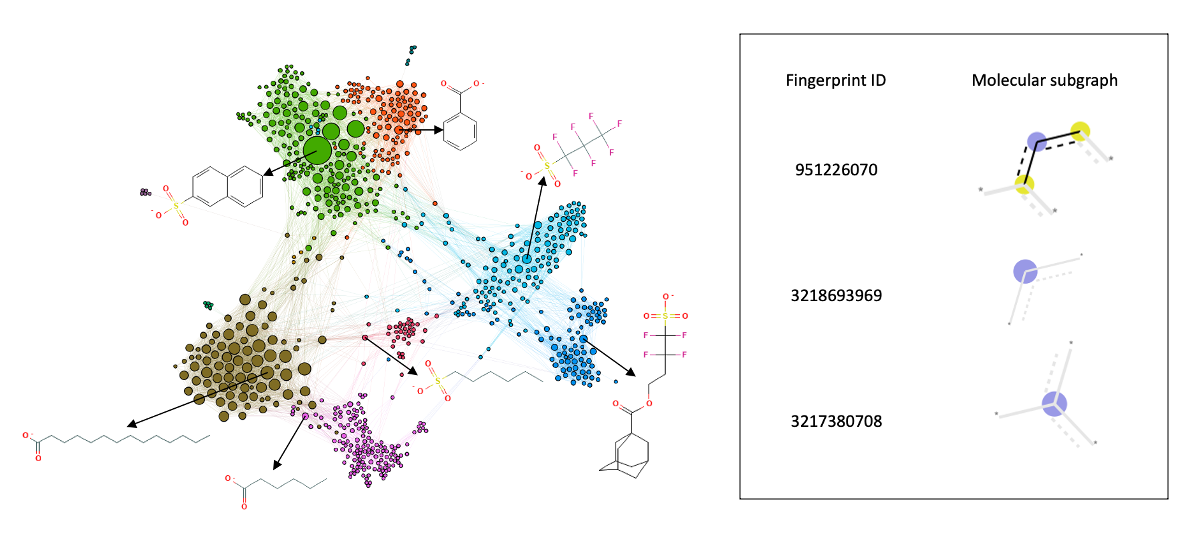}}

    \caption{Visualization of the CIRCA dataset and some of the structural fingerprints relevant to MPEGO analysis (inset). The CIRCA dataset is represented as a similarity network, where nodes correspond to anions and links connect nodes if Dice similarity of the respective anions is at least 0.5. Node colors encode clustering recovered via modularity analysis and node size encodes the number of connections of the node. Chemical structures are included for the most connected nodes in the main clusters. }\label{fig:CIRCA_Visualizations}
\end{figure}

\subsection{Generative Models}
{We utilized multiple generative models  for our validation. Particularly, Graph Convolutional Policy Network (GCPN)~\cite{gcpn2018you} and a Flow-based Autoregressive (GraphAF)~\cite{graphaf2020shi} were trained separately on ZINC-250K and MOSES datasets. Subsequently, $10,000$ valid molecules were generated from each model. We rely on GT4SD~\cite{manica2022gt4sd} for model implementations (experimental set-ups shown in Appendix B). 
Similarly, from the MOSES-trained GCPN and GraphAF, $14665$ and $1680$ molecules were generated respectively. 
For the CIRCA data, we employed MolGX~\cite{takeda2020molecular} and the Regression Transformer~\cite{born2022regression}) and generated 5000 new anions with MolGX and 32617 new anions with Transformer. Note that balanced number of samples were selected across datasets during our experimentation, i.e., $10000$, $1680$ and $5000$ samples were randomly selected for ZINC-250K-based, MOSES-based, CIRCA-based evaluations. }

\subsection{Feature Extraction and Preprocessing}
{We have extracted the following six features from SMILES representations of molecules in ZINK-250K and MOSES datasets and their corresponding generated datasets from GCPN and GrpahAF. The following features are selected for the evaluation: \textit{Aromaticity, ESOL, LogP, Weight, QED, SCScore}. 
On the other hand, we extracted fingerprints (see Fig.~\ref{fig:CIRCA_Visualizations}) that reflect structural details from CIRCA dataset and MolGX and Transformer generated datasets. Full details of features can be found in the Appendix C of the Supplementary Material. In cases when discretization of continuous features is required to compute the SIS values, we employ Yule's Y coefficient as our default objective measure as it satisfies multiple key requirements~\cite{tan2004selecting}. We also employ equal frequency discretization type as it better handles outliers in the data, with five bins for ZINC and with three bins for CIRCA based evaluations. While aggregating features for SAFIS and GAFIS computations using normal averaging. Note that  the proposed approach is flexible to utilize different discretization types, weighting strategies, and objective measures.}

\subsection{Evaluation Metrics}
 Our evaluation metrics include HIE's SIS, FIS, SAFIS, and GAFIS values that quantify the generation independence of models. The independence score is obtained  by normalizing the objective measure values  to $[0,1]$. Note  independence score $=1.0$ represents complete independence. 
We also utilize the histogram of features to provide a qualitative comparison.
The characterization of generation frequency analysis involves using the logical combination of  feature values to describe the identified subgroup, the size of the subgroup $N_s$, the odds ratio between $\mathcal{S}$ and $\widetilde{\mathcal{S}} = \mathcal{D} - \mathcal{S}$,  $95\%$ Confidence Interval (CI) and empirical $p$ value. We also reported the divergence score of the identified group from the expectation  in GFA and elapsed time to identify the group. All the experiments are conducted on a desktop machine, 2.9 GHz Quad-Core Intel Core i7 (processor), and 16 GB 2133 MHz LPDDR3 (memory).

\section{Results and Discussion}\label{sec:results_discussion}

\subsection{Hierarchical Independence Evaluation}~\label{subsec:nom_results}

     {Table~\ref{tab:hie_zinc250} provides extended FIS values across each of the features considered for comparing GCPN and GraphAF models with the training ZINC-250K  ($\mathcal{G} vs. \mathcal{T}$) and between each other - head-to-head ($\mathcal{G} vs. \mathcal{G}$). GCPN's GAFIS value of $0.882$, demonstrating competitive  generation independence  with GraphAF with GAFIS $=0.821$. This is further shown with their  head-to-head comparison, with Aromaticity ($FIS=1.0$) and LogP ($FIS=0.906$) features.  Divergent characteristics are also  demonstrated when the two models were evaluated based on QED, Weight and ESOL features, achieving FIS of $0.853$, $0.809$, and $0.662$, respectively in their head-to-head comparison. 
   SAFIS values are shown as an example of aggregation synthetic metrics, QED and LogP, where GCPN model achieves superiority compared to GraphAF. GAFIS values in the bottom row are derived from global aggregation of FIS values above.   Overall, the results show the flexibility of the MPEGO framework to evaluate models across different levels of feature interactions and comparison baselines, i.e., training data or another generated datasets.}

\begin{figure}[t]
    \centering
    \subfloat[]{ 
    \resizebox{0.7\linewidth}{!}{
    \begin{tabular}{llcccc}
 \toprule
      && \multicolumn{2}{c}{$\mathcal{G}$ vs. $\mathcal{T}$} &  \multicolumn{1}{c}{$\mathcal{G}$ vs. $\mathcal{G}$} \\
    & & \multicolumn{1}{c}{GCPN vs.} &   \multicolumn{1}{c}{GraphAF vs.} &\multicolumn{1}{c}{GCPN vs. } \\
        Level& {Weight Range} &  ZINC  & ZINC &  GraphAF \\
\midrule
    \rowcolor[HTML]{EFEFEF}  & $<$201.63           &         0.406 &            0.322 &            0.859 \\
& [201.63 , 258.1)  &         0.631 &            0.840 &            0.778 \\
   \rowcolor[HTML]{EFEFEF} SIS  & [258.1 , 308.39)  &         0.986 &            0.824 &            0.810 \\
& [308.39 , 361.47) &         0.671 &            0.637 &            0.962 \\
   \rowcolor[HTML]{EFEFEF}  & $\geq$361.47          &         0.586 &            0.939 &            0.638 \\
\bottomrule
\end{tabular}}}\label{tab:sis_weight}

    \subfloat[]{\includegraphics[width=0.7\linewidth]{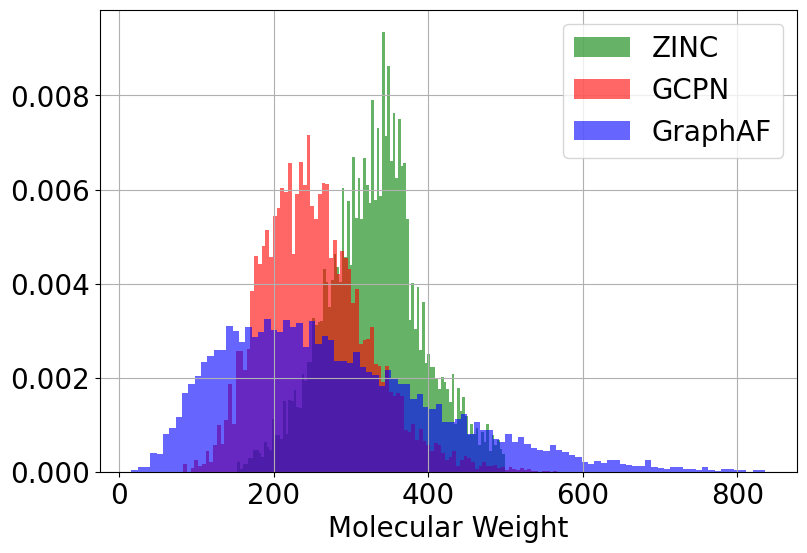}}\label{fig:hist_weight}
    \caption{(a) Example of Molecular Weight-based SIS values that provide sub-feature level evaluation of GCPN and GraphAF models trained on ZINC-250K; (b) histogram densities to provide qualitative visualization of the SIS values in (a)}\label{fig:sis_compare_models}
\end{figure}

 Furthermore, Fig.~\ref{fig:sis_compare_models} demonstrates the benefits of sub-feature level  evaluation (SIS) of GCPN and GraphAF models trained on ZINC-250K, using Molecular Weight as a feature example. Five ranges of the molecular weight in Fig.~\ref{fig:sis_compare_models} (a)
resulted from the discretization of the feature necessary for SIS evaluations. Similarly to Table~\ref{tab:hie_zinc250}, the comparison is performed between generative models and with the training dataset. Results demonstrate that GCPN and GraphAF generated molecules with similar weight characteristics but distinctively different at a few particular ranges, i.e., GCPN generated molecules similar to  ZINC-250K with range $[258.1,308.39)$ whereas GraphAF showed better resemblance with ZINC-250K at ranges $[201.63,258.1)$ and $\geq361.47$ Daltons. Divergent sub-level generation characteristics is also encoded by lower $\mathcal{G}$ vs. $\mathcal{G}$ independence score at those ranges. The histogram plot in Fig.~\ref{fig:sis_compare_models} (b) qualitatively compliments the insights from SIS values in Fig.~\ref{fig:sis_compare_models} (a) , where GraphAF is shown to generate more molecules with extreme weight values. Note that such  sub-feature level insights in Fig.~\ref{fig:sis_compare_models} are unique to our proposed framework as they are currently limited in the state-of-the-art of generative model evaluation. 

The results of comparisons of GCPN and GraphAF trained on MOSES dataset (Table~\ref{tab:hie_moses}) and MolGX and RT trained on CIRCA dataset (Table~\ref{tab:hie_circa}) demonstrate MPEGO's capabilities further. Comparisons between Table~\ref{tab:hie_zinc250} with Table~\ref{tab:hie_moses} show while MOSES is used as a training, GraphAF achieved more independence than when ZINC-250K is used, compared GCPN. 
Between MolGX and the RT, trained on CIRCA, molecules generated from the RT achieved a higher resemblance with CIRCA than MolGX. In their head-to-head comparison, MolGX and RT generated mostly divergent structural details, as it is demonstrated by very low FIS values for the majority of the fingerprints in $\mathcal{G}$ vs. $\mathcal{G}$ column of Table~\ref{tab:hie_circa}.

\begin{table}[t]
\captionsetup{font=footnotesize}
    \centering
    \caption{HIE scores for GCPN and GraphAF models trained on ZINC-250K dataset}\label{tab:hie_zinc250}
    \resizebox{0.7\linewidth}{!}{
    \begin{tabular}{llcccc}
    \toprule
      && \multicolumn{2}{c}{$\mathcal{G}$ vs. $\mathcal{T}$} &  \multicolumn{1}{c}{$\mathcal{G}$ vs. $\mathcal{G}$} \\
    & & \multicolumn{1}{c}{GCPN vs.} &   \multicolumn{1}{c}{GraphAF vs.} &\multicolumn{1}{c}{GCPN vs. } \\
        Level& {} &  ZINC  & ZINC &  GraphAF \\
        \midrule
        \rowcolor[HTML]{EFEFEF}  & Aromaticity      &         1.000 &            1.000 &            1.000 \\
       & ESOL             &         0.912 &            0.856 &            0.853 \\
      \rowcolor[HTML]{EFEFEF}& LogP            &         0.951 &            0.942 &            0.906 \\
     FIS & Weight &         0.656 &            0.713 &            0.809 \\
        \rowcolor[HTML]{EFEFEF} & QED              &         0.946 &            0.666 &            0.662 \\
        & SCScore          &         0.825 &            0.752 &            0.866 \\
        \midrule
        \rowcolor[HTML]{EFEFEF} SAFIS & QED+LogP &                 0.949 &	           0.804 &	0.784 \\ \midrule
GAFIS&  &         0.882 &            0.821 &             0.850 \\
        \bottomrule
    \end{tabular}

    }

\end{table}

\begin{table}[t]
    \centering
    \caption{HIE Scores for GCPN and GraphAF models trained on MOSES dataset}\label{tab:hie_moses}
    \resizebox{0.7\linewidth}{!}{
    \begin{tabular}{llcccc}
    \toprule
      && \multicolumn{2}{c}{$\mathcal{G}$ vs. $\mathcal{T}$} &  \multicolumn{1}{c}{$\mathcal{G}$ vs. $\mathcal{G}$} \\
    & & \multicolumn{1}{c}{GCPN vs.} &   \multicolumn{1}{c}{GraphAF vs.} &\multicolumn{1}{c}{GCPN vs. } \\
        Level& {} &  ZINC  & ZINC &  GraphAF \\
        \midrule
        \rowcolor[HTML]{EFEFEF}  & Aromaticity      &          1.000 &             1.000 &            1.000  \\
       & ESOL &             0.419 &             0.672 &            0.491 \\
      \rowcolor[HTML]{EFEFEF}& LogP &            0.219 &             0.682 &            0.319 \\
     FIS & Weight &         0.550 &             0.491 &            0.760 \\
        \rowcolor[HTML]{EFEFEF} & QED &             0.471 &             0.576 &            0.735 \\
        & SCScore          &         0.846 &             0.865 &            0.744 \\
        \midrule
        \rowcolor[HTML]{EFEFEF} SAFIS & QED+LogP &                0.345 &	0.629 &	0.527 \\ \midrule
GAFIS&  &        0.584 & 	0.714 &	0.675 \\

        \bottomrule
    \end{tabular}

    }
\end{table}


\begin{table}[!htb]
\caption{HIE Scores for MolGX and Regression Transformer (RT) models trained on CIRCA dataset
}
    \centering
    \resizebox{0.7\linewidth}{!}{
    \begin{tabular}{llcccc}
    \toprule
      && \multicolumn{2}{c}{$\mathcal{G}$ vs. $\mathcal{T}$} &  \multicolumn{1}{c}{$\mathcal{G}$ vs. $\mathcal{G}$} \\
    & & \multicolumn{1}{c}{MolGX vs.} &   \multicolumn{1}{c}{Transformer vs.} &\multicolumn{1}{c}{MolGX vs. } \\
        Level& Fingerprint &  MOSES  & MOSES &  Transformer \\
\midrule

 \rowcolor[HTML]{EFEFEF}  & 951226070  &           1.000 &                 1.000 &                 1.000 \\
 & 3218693969 &           0.599 &                 0.876 &                 0.500 \\
 \rowcolor[HTML]{EFEFEF}  & 2968968094 &           1.000 &                 1.000 &                 1.000 \\
 FIS & 882399112  &           1.000 &                 1.000 &                 1.000 \\
 \rowcolor[HTML]{EFEFEF}  & 2245384272 &           0.998 &                 0.847 &                 0.845 \\
 & 2246703798 &           1.000 &                 1.000 &                 1.000 \\
 \rowcolor[HTML]{EFEFEF}  & 3217380708 &           0.411 &                 0.661 &                 0.226 \\ \midrule
GAFIS& & 0.858 &  	0.912  & 	0.796\\

\bottomrule
\end{tabular}
}
\label{tab:hie_circa}
\end{table}

\subsection{Generation Frequency Analysis}\label{subsec:gfa_results}
{Table~\ref{tab:gfa_analysis} shows the divergence generation frequency of models, compared to the training or other generated set. For example, the GCPN model generated molecules with higher LogP ($\geq1.31$) and lower weight ($<258.1$ Daltons) compared to the training ZINC-250K. When we compare the two models head-to-head, GCPN tends to generate molecules with higher QED ($\geq 0.64$) values compared to GraphAF ($<0.5$). When MOSES data is used as validation, LogP values become the differentiation factor as GraphAF tends to generate lower values ($<3.91$) compared to GCPN ($\geq 3.91$). On CIRCA dataset, MolGX generates samples with a higher occurrence of '3217380708' fingerprint compared to the RT model and the training CIRCA. The size of the identified group, along with the multiplicative factor (q) and the odds ratio values, confirm the significance of the identified divergent generation in our generation frequency analysis.}

\begin{table*}[t]
    \centering
  \caption{Results from generation frequency analysis, validated across three datasets: ZINC\-250K (with GCPN and GraphAF models), MOSES (with GCPN and GraphAF models), and CIRCA (with MolGX and RT). We also compared the models head\-to\-head. The $q$ factor represents the multiplicative  increase in the odds of the outcome in the identified group compared to the null hypothesis. Elapsed  time is in seconds.}\label{tab:gfa_analysis}
    \resizebox{0.99\linewidth}{!}{
\begin{tabular}{lccp{3cm}cccccccc}
\toprule
         Model  &   Baseline &  Expected &                                             Subset & Group size &  Observed &  q factor &  Odds ratio &           95\% CI &  p\-value &     Score &  Elapsed \\
\midrule
  
    \rowcolor[HTML]{EFEFEF}  \multirow{2}{*}{GCPN}  &         ZINC &       0.5 &  LogP$\geq1.31$ \& Weight $<258.1$  &       5228 &     0.888 &     7.952 &   13.98 &  (12.75, 15.33) &      0.0 &  1781.619 &     9.21 \\    
     & \\
      &          GraphAF &       0.5 & SCScore $<3.54$ \& QED $\geq0.64$ \& Weight $<362.48$  &       8040 &     0.797 &     3.923 &        9.14 &    (8.54, 9.77) &      0.0 &  1493.252 &    10.99 \\   \cline{2-12} 
      & \\
      \rowcolor[HTML]{EFEFEF}   \multirow{2}{*}{GraphAF}         &   ZINC &        0.5 &  SCScore $\geq2.62$ \& QED $<0.64$ \& Weight $<258.1$ &       4072 &     0.946 &    17.425 &       27.71 &  (24.11, 31.85) &      0.0 &  1949.670 &     8.96 \\   & \\

    &        GCPN &       0.5 &  QED $<0.5$ &       5135 &     0.878 &     7.164 &       12.22 &  (11.17, 13.37) &      0.0 &  1647.798 &     7.73 \\

\midrule  & \\

 \rowcolor[HTML]{EFEFEF}   \multirow{2}{*}{GCPN}       &     MOSES &       0.5 &                QED $<0.5$ \&  $2.36\leq$LogP$<3.91$&       1422 &     0.985 &    66.714 &      397.89 &   (253.9, 623.53) &      0.0 &  870.290 &     4.00 \\  & \\

  & GraphAF &       0.5 &    LogP $\geq3.91$ \& Weight $<298.32$ &       1284 &     0.992 &   127.400 &      525.01 &  (279.16, 987.37) &      0.0 &  825.489 &     3.43 \\ \cline{2-12}  & \\

\rowcolor[HTML]{EFEFEF}  \multirow{2}{*}{GraphAF}  &           MOSES &       0.5 &        Weight $<250.38$  \& $-4.75\leq$ ESOL$<-3.59$ &        915 &     0.991 &   113.375 &      245.40 &  (121.72, 494.77) &      0.0 &  584.349 &     4.21 \\  & \\

  &         GCPN &       0.5 &         LogP $<3.91$ \& ESOL $<-4.75$ OR ESOL $\geq -3.59$ \& Weight $<250.38$ &       1297 &     0.976 &    40.839 &      162.96 &  (112.31, 236.46) &      0.0 &  744.637 &     3.33 \\

\midrule

  \rowcolor[HTML]{EFEFEF}  \multirow{2}{*}{RT}   &          CIRCA &       0.5 &                       2245384272 $\geq 1.0$ &        603 &     0.561 &     1.275 &        1.85 &      (1.43, 2.39) &      0.0 &    2.430 &     1.92 \\
     &         MolGX &       0.5 &  3217380708 $<4.0$ \& 3218693969 $<2.0$ &        353 &     0.989 &    87.250 &      286.60 &  (105.19, 780.82) &      0.0 &  218.783 &     2.27 \\ \cline{2-12}  
     
        \rowcolor[HTML]{EFEFEF}   \multirow{2}{*}{MolGX}   &               CIRCA &       0.5 &                     3217380708 $\geq 4.0$ &        484 &     0.804 &     4.095 &       14.94 &    (10.99, 20.31) &      0.0 &   93.805 &     2.30 \\ 
          &               RT &       0.5 &                       3217380708 $\geq 4.0$ &        416 &     0.935 &    14.407 &       61.39 &    (39.48, 95.47) &      0.0 &  186.404 &     2.11 \\ 
          \bottomrule
\end{tabular}
}
\end{table*}

{\subsection{Ablation Study}}

\begin{figure}[t]
    \centering
    \subfloat[]{\includegraphics[width=0.495\linewidth]{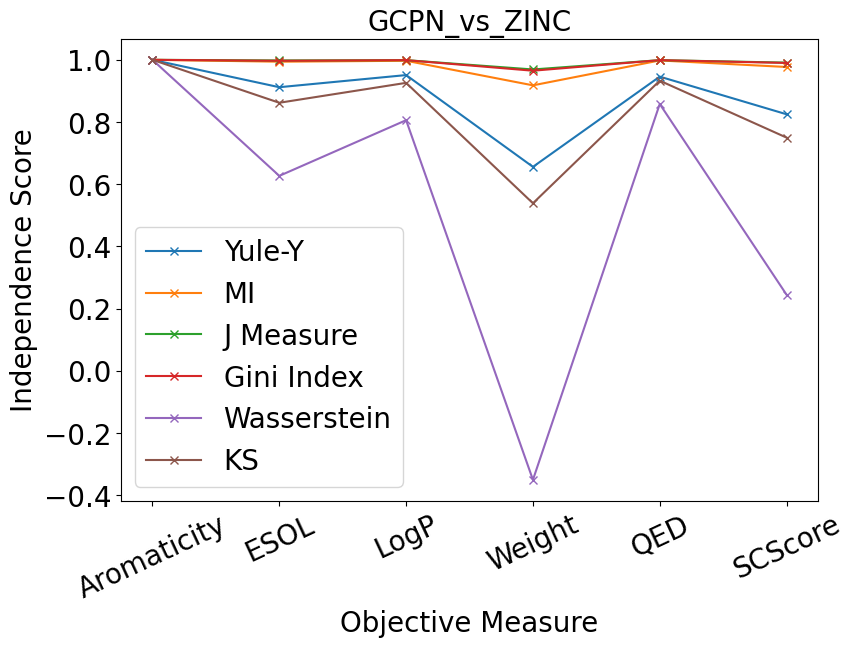}}\label{fig:om_gcpn_training}
    \subfloat[]{\includegraphics[width=0.495\linewidth]{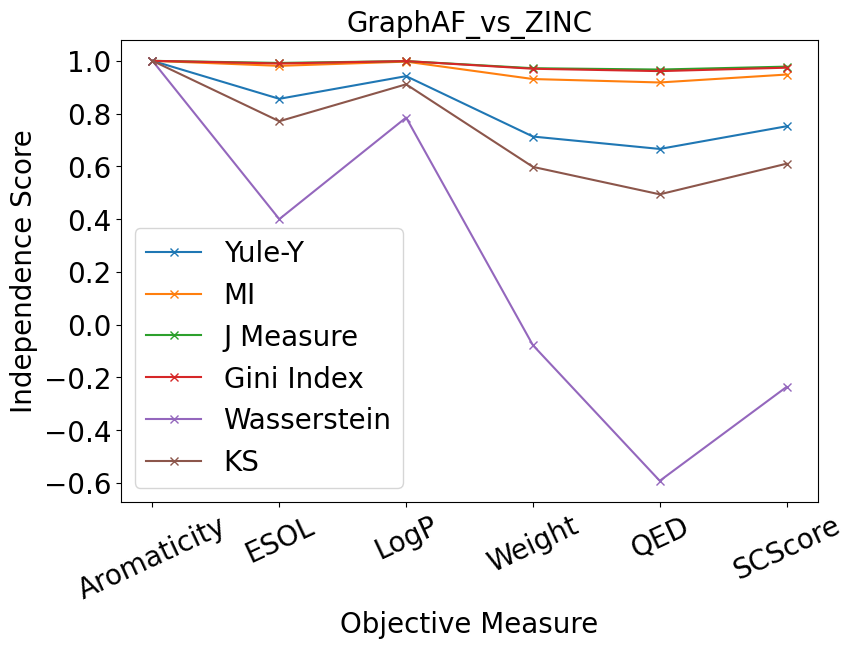}}\label{fig:om_graphaf_training}
    \caption{Different objective measures provided similar ranking per feature-level performance score (FIS) of the generative models }\label{fig:ablation_objective_measures}
\end{figure}

\begin{figure}[t]
    \centering
    \subfloat[]{\includegraphics[width=0.495\linewidth]{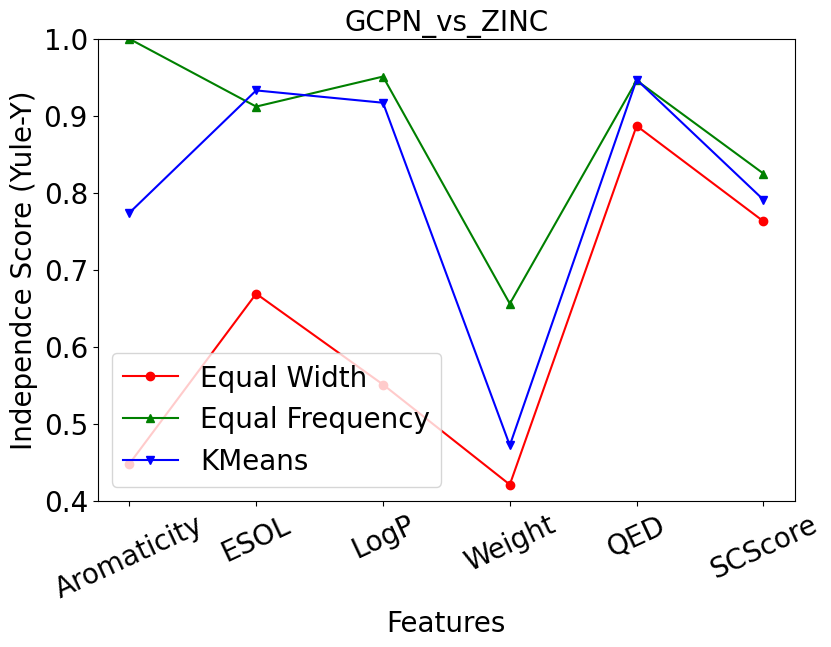}}\label{fig:dt_gcpn_training}
    \subfloat[]{\includegraphics[width=0.495\linewidth]{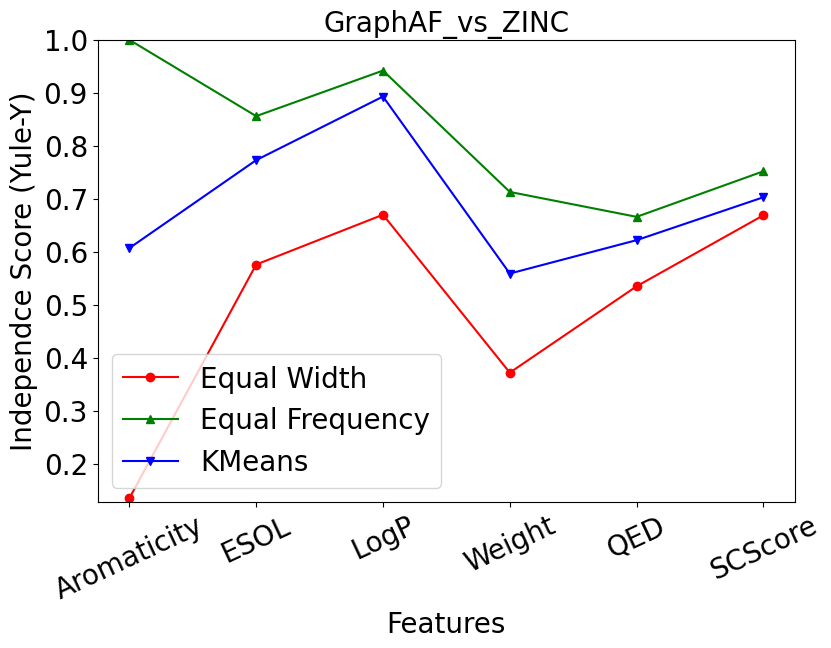}}\label{fig:dt_graphaf_training}
    \caption{Different discretization types provided similar ranking per feature-level performance score (FIS) of the generative models }\label{fig:ablation_discretization_types}
\end{figure}

{We conducted ablation studies to validate the robustness of MPEGO for different design choices of objective measures (see Fig.~\ref{fig:ablation_objective_measures}) and discretization types (see Fig.~\ref{fig:ablation_discretization_types}).  Though objective measures could vary in whether they require discretization or not, the similar shapes of the plots in Fig.~\ref{fig:ablation_objective_measures} demonstrate similar ranking performance across features. For example, in GCPN vs. ZINC comparison (Fig.~\ref{fig:ablation_objective_measures} (a)), Molecular weight achieved the least score among features under all the different objective measures employed.  The same is true for QED in the GraphAF vs. ZINC validation in   Fig.~\ref{fig:ablation_objective_measures} (b).
We also validated the impact of the discretization types as shown in Fig.\ref{fig:ablation_discretization_types} (a) and (b). In both cases,  \emph{equal width} and \emph{$k$-means} discretizations provide similar patterns while \emph{equal frequency} discretization demonstrated a slightly different pattern, particularly for features with skewed distribution as \emph Aromaticity. Overall, all three types of discretization provide competitive performance scores, thereby validating the stability of our MPEGO framework. More ablation results are in Appendix E of the Supplementary Material.
}

\section{Conclusion and Future Work}\label{sec:conclusion}

 We proposed MPEGO - a simple, generalizable, and model-agnostic  evaluation framework of generative models validated for material discovery domain.  
 MPEGO consists of two main performance evaluation blocks: Hierarchical Independence Evaluation (HIE) and Generation Frequency Analysis (GFA). HIE follows a bottom-up approach to quantify the generation performance of a model, starting from per sub-feature level (at the bottom) to the global aggregation of features (at the top). Thus, HIE provides a flexible performance evaluation of generative models. {Particularly by evaluating the generation independence of models compared with the training data or other generative models using an objective measure set by a user. GFA is applied to detect and characterize divergent generation characteristics. Different from the existing evaluation platforms, MPEGO provides interpretable insights that aim to facilitate interactions with subject matter experts,  which is crucial to develop Trustworthy AI solutions.} 
{The proposed MPEGO toolkit was validated with multiple datasets (ZINC-250K, MOSES, and CIRCA) and generative models trained on these datasets, including GCPN, GraphAF, MolGX, and RT. Conditioned on the training samples in these datasets,  GCPN, GraphAF, and RT achieved higher generation independence, compared to their counterparts, GraphAF, GCPN, and MolGX models, respectively.} 
{Future work aims to evaluate generative models from different domains to further validate the domain-agnostic nature of the MPEGO framework. We also plan to utilize MPEGO to improve the efficiency of latent-based analyses, such as creativity characterization~\cite{cintas2022towards} and out-of-distribution detection~\cite{kim2022out}. 
MPEGO will be natively integrated into the GT4SD, the Generative Toolkit for Scientific Discovery~\cite{manica2022gt4sd} and the source code and experiments will be available at:~\url{https://github.com/GT4SD/mpego}.}

\bibliographystyle{unsrt}  
\bibliography{mpego}  
\newpage
\section*{Appendix A: Steps to prepare CIRCA dataset}

Initially, we carried out a patent search in CIRCA for ionic photoacid generators (PAGs) that are sulfonium or iodonium salts. We extracted anionic components of the sulfonium and iodonium salts, finding 493 unique anions. We ran additional CIRCA search to enrich the dataset with anions relevant to photolithography using the following query: ``PAG'' OR ``photo-acid generator'' OR ``photo initiator'' OR ``acid generating agent'' OR ``photocationic initiator'' OR ``photocationic polymerization initiator'' OR ``cationic initiator'' OR ``cationic type initiator'' The additional search returned ionic compounds with 1398 anions, 189 of these 1398 anions were among the initial 493 anions.
We processed the set of collected anions, keeping only anions with charge -1 and more than 1 carbon atom, reducing the dataset to 1086 anions. Out of 1086 anions, we were able to evaluate environmental and toxicological properties of 866 anions, that comprise the dataset used in this study

\section*{Appendix B: Experimental Setup for GCPN and GraphAF training}

 \begin{table}[ht]
     \centering
    \caption{Summary of hyper-parameters set-up to train the two graph-based models: GCPN and GraphAF using  ZINC-250K dataset.}
     \label{tab:generation_setup}
     \resizebox{0.7\linewidth}{!}{
     \begin{tabular}{l|c|c}
     \toprule
    & \multicolumn{2}{c}{Models} \\
     Setting  & GCPN & GraphAF \\ \midrule
    Input Dimension &  18 & 9 \\ \hline  
    Number of relation & 3 & 3  \\ \hline     
    Batch normalization &  False &  True \\ \hline
    Atom types &  [6-9, 15-17, 35, 53] & [6-9, 15-17, 35, 53]  \\ \hline
    Hidden dimensions &  [256, 256, 256, 256] & [ 256, 256, 256] \\ \hline
     \end{tabular}}

 \end{table}

\section*{Appendix C: Descriptions of features}

 \begin{table}[ht]
    \centering
    \caption{Features extracted from samples generated from GCPN and GraphAF models trained on ZINC-250k and MOSES datasets.}
    \label{tab:features}
    \resizebox{0.7\linewidth}{!}{
    \begin{tabular}{l|p{8cm}}
  \toprule
  \textbf{Feature}  &\textbf{Description} \\ \midrule
    \multirow{1}{*}{QED} &  Qualitative Estimate of Drug-likeness \\ \hline
    ESOL &  Estimated SOLubility  \\ \hline
    SCScore &  Synthetic Complexity Score  \\ \hline
    SAS &  Synthetic Accessibility Score.  \\ \hline
    Scaffold & A molecule is identical to its scaffold \\ \hline
    Ring & A molecule contains a ring structure \\ \hline
    LargeRing &  A molecule contains a ring structure with more than $6$ atoms  \\ \hline
    Aromatic &  A molecule contains an aromatic ring \\ \hline
    SteroCharacter &   SMILES string contains a stereochemistry string. \\ \hline
    Stereocenter &  Whether a molecule contains a stereocenter, \\ \hline
    Heterocycle &  A molecule has at least one ring with at least two different atoms \\ \hline
    Lipinski &  A molecule adheres to the Lipinski's rule of five. \\ \hline
    HBondDonor &  A molecule has not more than 5 hydrogen bond donors \\ \hline
    HBondAcceptor &  A molecule has not more than 10 hydrogen bond acceptors \\ \hline
    MolecularWeight &  The molecular mass  in  Daltons \\ \hline
  LogP  &  Logarithmic partition coefficient  \\ \bottomrule
    \end{tabular}
    }

\end{table}

Multiple fingerprints, informing the presence of  particular structural elements in a given molecule, are extracted from CIRCA dataset and samples generated from MolGX and RT models. The following fingerprints are selected for evaluation in this study due to their higher variance across the datasets.

\begin{table}[ht]
    \centering
    \begin{tabular}{M{2.5cm}M{2.5cm}}
    \toprule
    Fingerprint& Molecular Subgraph \\ \midrule
      {2245384272}  &     { \includegraphics[width=0.5\linewidth]{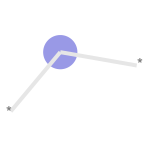}} \\ \hline

      {951226070}  &     { \includegraphics[width=0.5\linewidth]{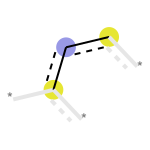}} \\ \hline

     {3217380708}  &     { \includegraphics[width=0.5\linewidth]{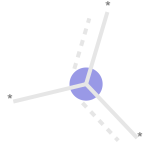}} \\ \hline
            
      {2246703798}  &     { \includegraphics[width=0.5\linewidth]{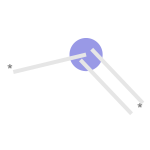}} \\ \hline
                  
    {882399112}  &     { \includegraphics[width=0.5\linewidth]{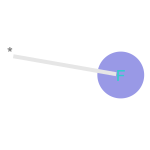}} \\ \hline
                        
  {3218693969}  &     { \includegraphics[width=0.5\linewidth]{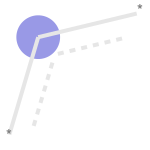}} \\ \hline
                              
{2968968094}  &     { \includegraphics[width=0.5\linewidth]{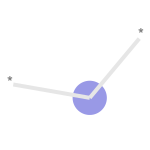}} \\ \bottomrule
    \end{tabular}
    \caption{Fingerprints (and their molecular subgraph visualizations) selected for evaluations in CIRCA datasets and its associated models: MolGX and RT.}\label{tab:my_label}
\end{table}

 \section*{Appendix D: List of Objective measures}
 The MPEGO framework could handle different objective measures to quantify sub-feature or feature-level-based association between generated and training datasets or between two generated datasets. The list of objective measures implemented in the MPEGO framework includes:\textit{ Odds ratio, Yule's Y Coefficients, Yule's Q coefficients, Mutual Information,J Measure, Gini Index, Wasserstein Distance, Kstest, Support, Confidence, Laplace, Collective Strength, Kappa, Conviction, Interest, Cosine, Piatetsky Shapiro, Certainty Factor, Added Value,  Jaccard and Kolmogorov–Smirnov.}
 
\newpage
\section*{Appendix E: Ablation results}
\begin{figure}[ht]
    \centering
    \subfloat[]{\includegraphics[width=0.495\linewidth]{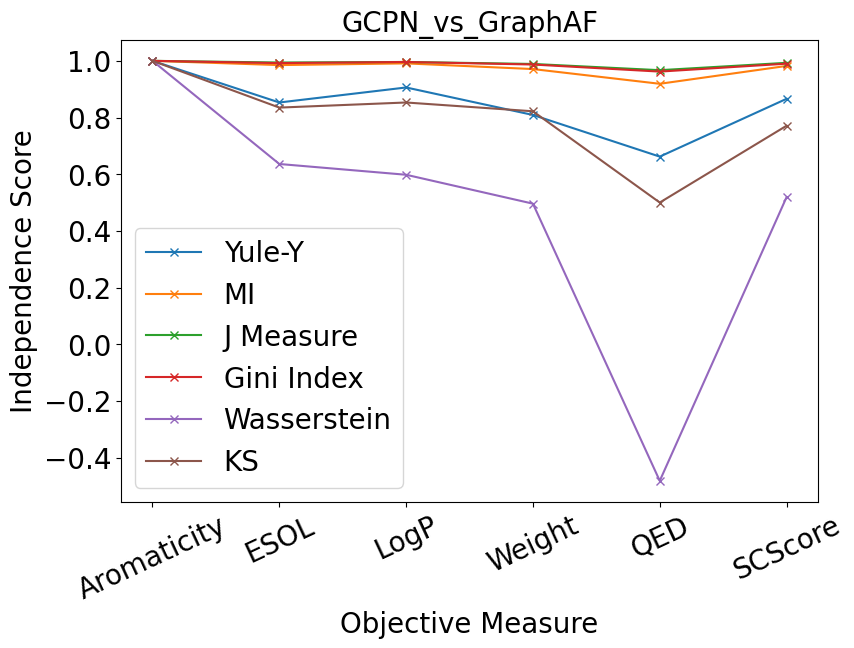}}
     \subfloat[]{\includegraphics[width=0.495\linewidth]{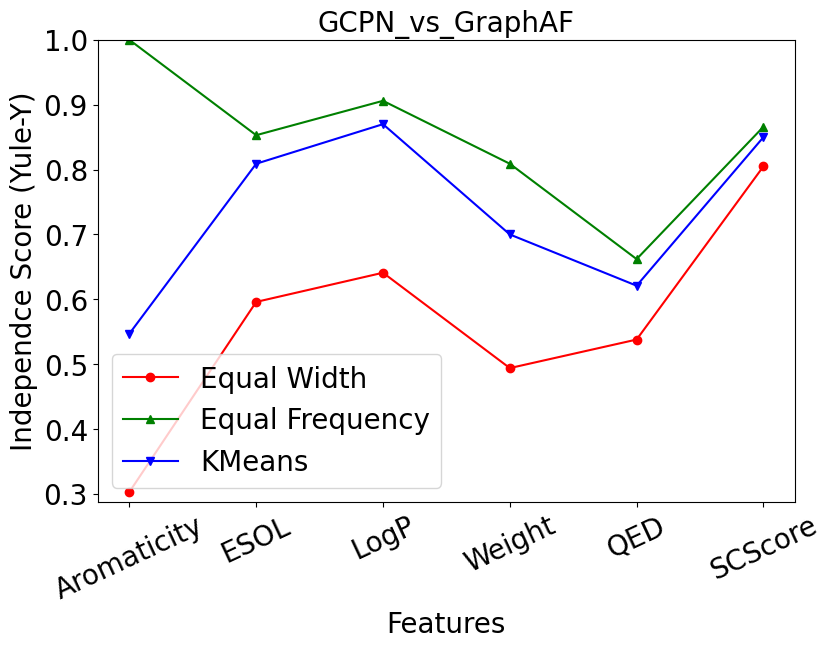}}
           \caption{Results from the ablation study on the variations of (a) objective measures and (b) discretization types on the FIS values, validated with a head-to-head comparison of GCPN and GraphAF models, trained on ZINC-250K.}\label{fig:ablation_head_to_head}
\end{figure}

Similar ablation studies are applied for head-to-head comparison of models as shown in Fig.~\ref{fig:ablation_head_to_head}, where the robustness of the MPEGO framework is demonstrated across varieties of objective measures (Fig.~\ref{fig:ablation_head_to_head} (a)) and discretization types (Fig.~\ref{fig:ablation_head_to_head} (b)).
\vspace{-5cm}
\end{document}